\documentclass[sigconf]{acmart}
\AtBeginDocument{%
  }

\copyrightyear{2026}
\acmYear{2026}
\setcopyright{cc}
\setcctype{by-nc-nd}
\acmConference[MM '26]{Proceedings of the 34th ACM International Conference on Multimedia}{November 10--14, 2026}{Rio de Janeiro, Brazil}
\acmBooktitle{Proceedings of the 34th ACM International Conference on Multimedia (MM '26), November 10--14, 2026, Rio de Janeiro, Brazil}
\acmDOI{10.1145/3767308.3836125}
\acmISBN{979-8-4007-2213-4/2026/11}
\usepackage{booktabs}
\usepackage{multirow}
\usepackage{makecell}
\usepackage[table]{xcolor}
\usepackage{graphicx}
\usepackage{amsmath,amssymb}
\usepackage{tabularx}
\usepackage{pifont}
\usepackage{balance}

\definecolor{formgreen}{rgb}{0.0,0.5,0.0}
\definecolor{bestgray}{gray}{0.92}
\definecolor{grayhighlight}{gray}{0.95}
\newcommand{\cmark}{\textcolor{formgreen}{\ding{51}}}
\newcommand{\xmark}{\textcolor{gray!60}{\ding{55}}}
\newcommand{\best}[1]{\cellcolor{bestgray}\textbf{#1}}

\begin{document}

\title{Flow-Map Distillation on Relation Manifolds for \\Image Restoration}

\author{Zihao He}
\orcid{0009-0004-6274-2365}
\email{hzh813009547@gmail.com}
\affiliation{%
  \department{School of Artificial Intelligence}
  \institution{Shanghai Jiao Tong University}
  \city{Shanghai City}
  \country{China}
}

\author{Songhua Liu}
\orcid{0000-0003-1033-5122}
\email{liusonghua@sjtu.edu.cn}
\affiliation{%
  \department{School of Artificial Intelligence}
  \institution{Shanghai Jiao Tong University}
  \city{Shanghai City}
  \country{China}
}

\renewcommand{\shortauthors}{Zihao He and Songhua Liu}

\begin{abstract}
Knowledge distillation for image restoration typically aligns 
intermediate features or relation matrices between teacher and 
student networks as static targets, ignoring the dynamic structure 
of the knowledge transfer process. In this paper, we propose 
\textbf{Flow-Map Distillation on Relation Manifolds (FoRM)}, which 
reformulates relation-based knowledge transfer as a continuous flow 
mapping problem on the relation manifold. Rather than regressing a 
constant velocity field between student and teacher relation states, 
FoRM learns a flow map operator $\mathcal{F}_\theta(\mathbf{z}, t, s)$ 
that directly predicts the relation state at any target time $s$ 
given the current state at time $t$, enabling richer 
trajectory-level supervision. To ensure global self-consistency 
of the learned flow map, we introduce a \emph{safe semigroup 
consistency} constraint that enforces compositional agreement 
using ground-truth bridge states, eliminating phantom-state error 
accumulation. An endpoint anchoring loss further prevents the 
operator from drifting away from the teacher target. Extensive 
experiments on five image restoration tasks, including 
super-resolution, deraining, denoising, deblurring, and 
low-light enhancement, demonstrate consistent gains over 
state-of-the-art distillation baselines across multiple 
backbone architectures, reducing training variance by 
approximately \textbf{50\%} compared to naive flow matching 
distillation while achieving superior restoration quality.
\end{abstract}

\ccsdesc[500]{Computing methodologies~Reconstruction}
\ccsdesc[300]{Computing methodologies~Neural networks}
\ccsdesc[300]{Computing methodologies~Image processing}

\keywords{knowledge distillation, super-resolution, flow matching, 
relation manifold, semigroup consistency}


\maketitle

\section{Introduction}
{\sloppy
Image restoration---encompassing super-resolution, deraining,
denoising, deblurring, and low-light enhancement---has become
increasingly critical in bandwidth-constrained and
edge-computing scenarios such as mobile imaging, medical
diagnosis, and satellite analysis~\cite{dong2014learning,
lim2017edsr, zhang2018rcan, zamir2022restormer}.
While recent advances have pushed restoration quality to
remarkable levels through deeper and wider
architectures~\cite{zhang2018rcan, lim2017edsr, zamir2022restormer},
these gains come at substantial computational cost,
making direct deployment on resource-limited devices
impractical.
Knowledge distillation~(KD)~\cite{logits}
offers a principled pathway to bridge this gap: by
transferring structural knowledge from a powerful teacher
network to a compact student, one can recover much of the
teacher's representational capacity without incurring its
inference overhead.
Consequently, KD has become a standard tool in lightweight
image restoration pipelines~\cite{he2020fakd,
gou2021knowledge,mansourian2025comprehensive}.

\begin{figure}[t]
  \centering
  \includegraphics[width=\linewidth,height=0.52\linewidth]{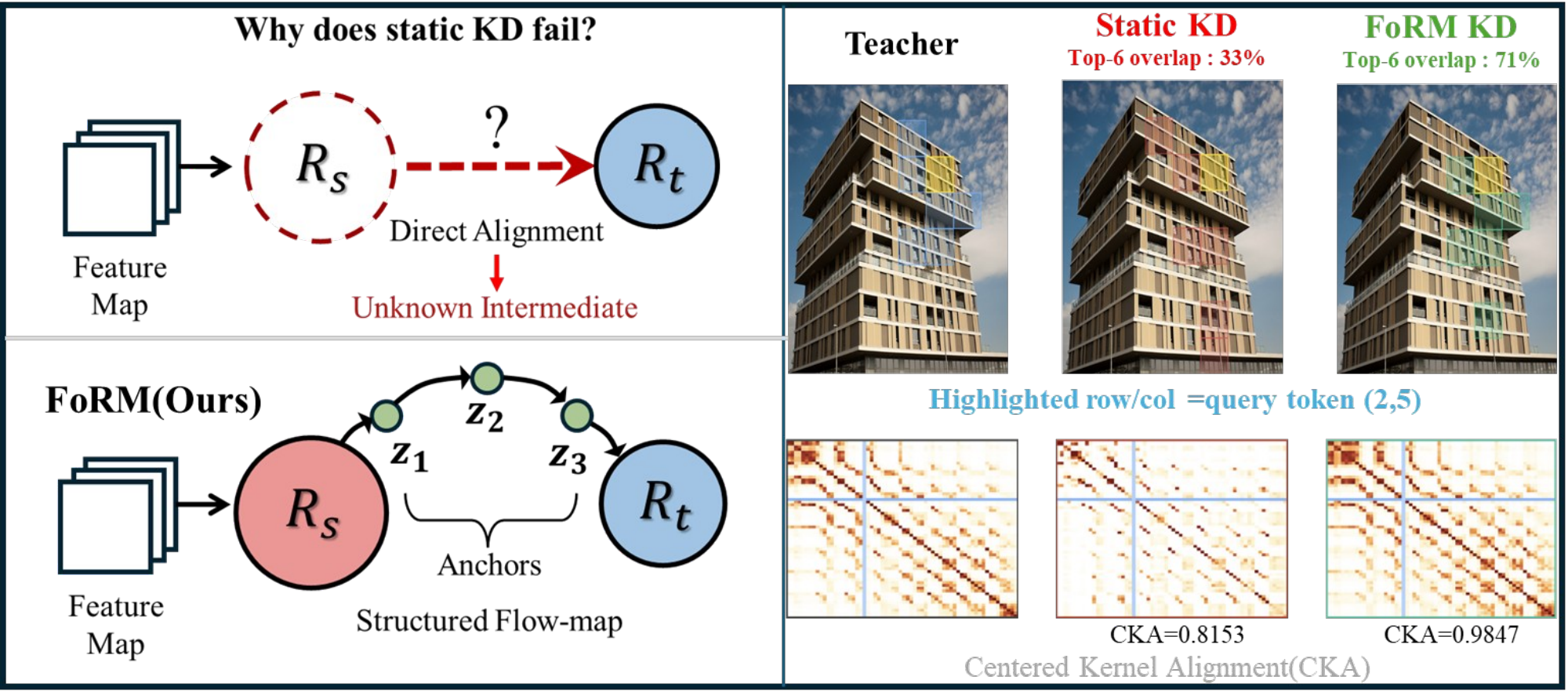}
  \Description{Left column: two schematic diagrams. The upper one, titled ``Why does
  static KD fail?'', connects a student relation state to a teacher relation
  state by a single dashed arrow labelled Direct Alignment with a question
  mark, marking the intermediate path as unknown. The lower one, titled FoRM
  (Ours), connects the same two states through three intermediate anchor
  states z1, z2 and z3 that form a structured flow map. Right column: a
  building photograph from Urban100 restored by the Teacher, by Static KD and
  by FoRM, each shown above its relation matrix rendered as a heat map with
  the row and column of one query token highlighted. Static KD reaches 33
  percent top-6 patch overlap and CKA 0.8153, whereas FoRM reaches 71 percent
  and CKA 0.9847.}
  \caption{
    \textbf{Motivation and overview of FoRM.}
    \textit{Left}: Static KD forces the student to directly
    match the teacher without intermediate guidance,
    leaving the optimisation path undefined.
    \textit{Middle}: FoRM constructs structured flow-map
    anchors $\mathbf{z}_1, \mathbf{z}_2, \mathbf{z}_3$
    along the distillation trajectory.
    \textit{Right}: Relation map comparison on Urban100.
    FoRM achieves significantly higher top-6 patch overlap
    (71\% vs.\ 33\%) and CKA (0.9847 vs.\ 0.8153) with
    the teacher, indicating better relational structure
    alignment.
  }
  \label{fig:teaser}
\end{figure}

Despite broad adoption, existing KD methods for image restoration
share a fundamental modelling assumption: distillation is treated
as a \emph{static alignment} problem. Whether matching intermediate
feature maps~\cite{romero2014fitnets}, attention
matrices~\cite{zagoruyko2017paying}, or pairwise relation
structures~\cite{park2019rkd}, the student reproduces a fixed
teacher target at each layer, typically via an $\ell_1$ or $\ell_2$
penalty.
This one-shot endpoint matching ignores the
\emph{trajectory} along which the student converges toward
the teacher: gradients from a static, potentially
over-constrained target can cause irregular curvature
throughout training, often manifesting as persistent
quality-metric oscillation in later epochs, premature
plateau, or suppression of fine-grained detail when the
teacher's representation lies far from the student's
capacity frontier.
A more complete problem formulation should therefore ask
not only \emph{where} the student should arrive, but
\emph{how} it should travel there.

A natural remedy is a \emph{dynamic, process-aware}
distillation objective. Flow matching~\cite{lipman2023flow,
liu2023flow, albergo2023stochastic} offers a principled
continuous-time framework: by constructing a probability path
between an initial and a target distribution, one can supervise
intermediate states rather than only the endpoint.
Under a straight-line bridge, applying this naively to KD amounts
to regressing a \emph{constant velocity field}
$\mathbf{u}^* = z_1 - z_0$---richer than static matching, yet
structurally deficient: a single velocity direction captures
neither the compositional structure of mappings at different
bridge points nor the global self-consistency a well-behaved flow
should satisfy. Enforcing semigroup consistency by recursively
feeding model-predicted states back into the network moreover
introduces \emph{phantom state error accumulation}, as predictions
drift from the true bridge and the outer mapping compounds the
deviation~\cite{boffi2025consistency, song2023consistency}.
These limitations motivate a more structured flow-based
formulation that captures the compositional geometry of
the relation space and avoids self-generated shortcut
solutions.

In this paper, we propose \textbf{Flow-Map Distillation
on Relation Manifolds~(FoRM)}, a novel knowledge
distillation framework for image restoration that
reformulates relation-based knowledge transfer as a
\emph{continuous flow mapping problem} on the relation
manifold.
The distillation state is defined as the vectorised
pairwise relation matrix
$z = \mathrm{vec}(\mathcal{R}(\phi))$,
where $\mathcal{R}$ computes a softmax-normalised token
affinity from pooled intermediate features.
The core of FoRM is a \emph{flow map operator}
$\mathcal{F}_\theta(\mathbf{z}, t, s)$ that directly
predicts the relation state at any target time $s$ given
the current state at time $t$ on the linear interpolation
bridge between student and
teacher~\cite{sabour2025alignflow, boffi2025consistency}.
To enforce global self-consistency, we introduce a
\emph{safe semigroup consistency} constraint: rather than
feeding back model-predicted intermediate states, both
branches are conditioned on \emph{ground-truth} bridge
points, eliminating phantom-state error accumulation
while preserving the compositional structure the
constraint is designed to enforce.
An endpoint anchoring loss further ensures the operator
remains anchored to the teacher target at $s=1$.
Together, these components yield richer trajectory-level
supervision and markedly more stable training dynamics.

The contributions of this work are threefold:
\begin{itemize}
    \item We recast relation-based image restoration KD
    as a flow map distillation problem on the relation
    manifold, shifting the paradigm from static endpoint
    alignment to dynamic trajectory-level knowledge
    transfer applicable across diverse restoration tasks.

    \item We propose safe semigroup consistency with
    endpoint anchoring---two structural constraints that
    enforce compositional self-consistency of the flow
    map without model-generated phantom states, reducing
    late-stage training variance by approximately 50\%
    compared to naive flow matching distillation.

    \item Extensive experiments across five image
    restoration tasks---super-resolution, deraining,
    denoising, deblurring, and low-light
    enhancement---demonstrate consistent gains over
    state-of-the-art distillation baselines across
    multiple backbone architectures, with ablation
    studies confirming the necessity of each structural
    component.
\end{itemize}

The remainder of this paper is organised as follows.
Section~\ref{sec:related} reviews related work on image
restoration, knowledge distillation, and flow-based
models.
Section~\ref{sec:method} details the FoRM framework.
Section~\ref{sec:experiments} presents experimental
results and ablation studies.
Section~\ref{sec:conclusion} concludes the paper, with
OT diagnostic analysis provided in the Appendix.
}

\begin{figure*}[t]
  \centering
  \includegraphics[width=\textwidth,height=0.295\textwidth]{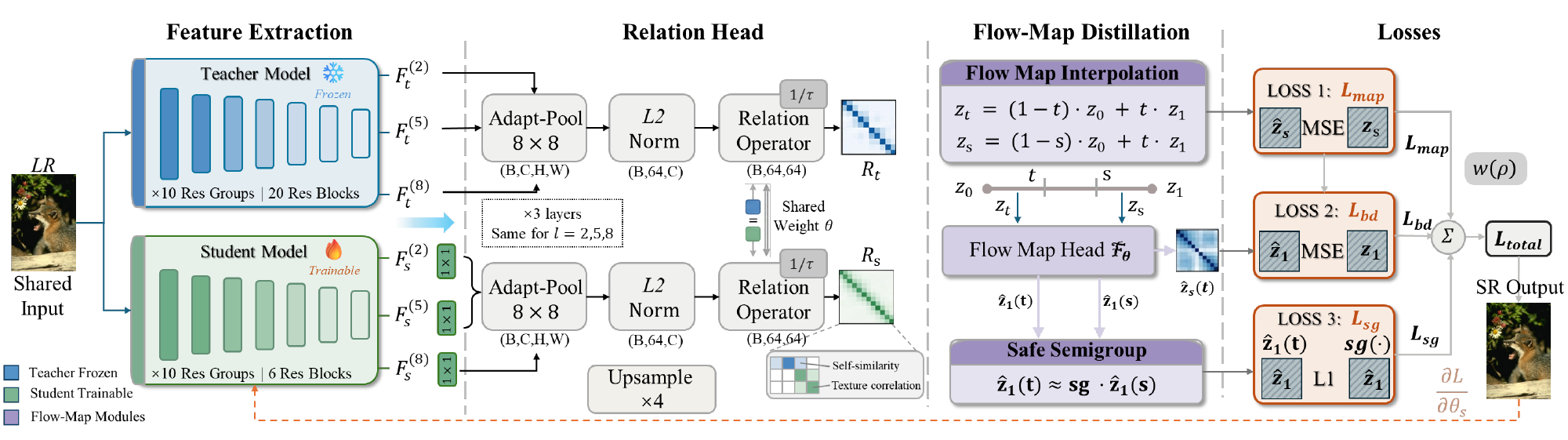}
  \Description{A four-stage pipeline diagram read from left to right. Feature Extraction: a
  shared low-resolution input feeds a frozen Teacher model of 20 residual
  blocks and a trainable Student model of 6 residual blocks, both emitting
  features at layers 2, 5 and 8. Relation Head: each feature passes through
  adaptive pooling to 8 by 8, L2 normalisation and a relation operator with
  shared weights, producing teacher and student relation matrices displayed as
  heat maps. Flow-Map Distillation: ground-truth bridge states are formed by
  linear interpolation between the student and teacher states, and a flow map
  head predicts relation states at arbitrary target times, with a safe
  semigroup branch that applies stop-gradient to one side. Losses: three
  blocks, map distillation under MSE, endpoint anchoring under MSE and
  semigroup consistency under L1, are combined through a schedule weight into
  the total objective that drives the super-resolved output.}
  \caption{
    \textbf{Overview of the FoRM framework.}
    Teacher and student features are extracted at layers
    $\ell \in \{2, 5, 8\}$, pooled to $8{\times}8$, and
    converted to softmax-normalised relation matrices
    $R_t$ and $R_s$ via a shared Relation Head.
    Ground-truth bridge states $z_t$, $z_s$ are constructed
    by linear interpolation between student and teacher
    relation states; the flow map head $\mathcal{F}_\theta$
    predicts relation states at arbitrary target times.
    Three losses, $\mathcal{L}_{\mathrm{map}}$,
    $\mathcal{L}_{\mathrm{bd}}$, and
    $\mathcal{L}_{\mathrm{sg}}$, jointly supervise
    $\mathcal{F}_\theta$ under schedule weight $w(\rho)$.
  }
  \label{fig:framework}
  \vspace{-4mm}
\end{figure*}

\section{Related Work}
\label{sec:related}

\subsection{Knowledge Distillation for Image Restoration}

Existing KD methods for image restoration share a common
paradigm: distillation as \emph{static target alignment}.
FitNets~\cite{romero2014fitnets} pioneered feature-level
matching via $\ell_2$ penalties on intermediate
activations, a strategy extended to attention
statistics~\cite{zagoruyko2017paying} and
restoration-specific feature affinity
objectives~\cite{he2020fakd}.
Across these variants, the teacher representation serves
as a \emph{fixed target}: the student is penalised for
deviating from it at each layer, and this one-shot
endpoint matching can manifest as late-stage oscillation,
premature plateau, or over-constrained suppression of
fine-grained detail.

Relation-based distillation~\cite{park2019rkd,
liu2019crd, tung2019similarity} addresses
cross-architecture brittleness by transferring pairwise
structural geometry rather than raw activations.
More recently, VRM~\cite{zhang2025vrm} demonstrated that
virtual-view augmentation further reduces spurious
relation matching.
However, all existing relation KD methods remain
fundamentally static: $R_t$ is a fixed per-forward-pass
target, and adaptive weighting
schemes~\cite{zhou2021distilling} offer heuristic relief
without addressing \emph{how} the student relation state
should evolve toward the teacher's.
Modelling the transfer process itself requires a
continuous-time framework, which we turn to next.

\subsection{Flow Matching and Flow-Map Consistency}
{\sloppy
Flow matching~\cite{lipman2023flow, liu2023flow,
albergo2023stochastic} replaces endpoint supervision with
continuous-time path objectives, training a velocity
field to steer samples along an interpolation bridge.
Under the straight-line bridge, the simplest instantiation
regresses a constant velocity $\mathbf{u}^* = z_1 - z_0$,
which is richer than static matching but remains limited:
it captures only net displacement, not the mapping
geometry across arbitrary time pairs $(t, s)$, and
single-sample conditional velocity targets exhibit high
variance across time steps that impedes
convergence~\cite{lee2026stable}.
In restoration KD, where the distillation signal competes
with the reconstruction objective, this instability is
especially costly.

These limitations motivate learning \emph{flow map
operators} $\mathcal{F}_\theta(z, t, s)$ that directly
predict the state at any target time $s$ from the state
at time $t$~\cite{sabour2025alignflow,
boffi2025consistency}, providing richer trajectory-level
supervision and naturally subsuming velocity regression
as a special case.
A key structural principle is the semigroup property
$\mathcal{F}(z,t,1) =
\mathcal{F}(\mathcal{F}(z,t,s),s,1)$,
but enforcing it by recursively feeding model-predicted
intermediate states introduces \emph{phantom state error
accumulation}~\cite{song2023consistency,
boffi2025consistency}, a risk amplified in KD where both
endpoints carry fixed semantic content.
In the restoration domain, generative methods such as
FlowSR~\cite{xu2025flowsr}, CTMSR~\cite{you2025ctmsr}, and
FluxSR~\cite{li2025fluxsr} have demonstrated flow-based trajectory
modelling for one-step SR, as does ultra-high-resolution
synthesis~\cite{freeswim, vibe}; all operate in the
generative sampling regime and do not address discriminative
restoration KD.
FoRM fills this gap by learning $\mathcal{F}_\theta$ on
the relation manifold with a \emph{safe semigroup}
constraint conditioned on ground-truth bridge points,
paired with endpoint anchoring to prevent manifold drift.
}

\section{Method}
\label{sec:method}

\subsection{Overview}

Given a frozen teacher $T(\cdot)$ and a trainable student
$S(\cdot)$, FoRM transfers relational knowledge by learning
a \emph{flow map operator} $\mathcal{F}_\theta(z, t, s)$
that predicts the relation state at any target time $s$
from any source time $t$ along a ground-truth linear bridge
between the student and teacher relation states.
Three complementary objectives---map distillation, endpoint
anchoring, and safe semigroup consistency---jointly supervise
the operator across the full trajectory, replacing the static
endpoint-alignment paradigm of prior image restoration KD
methods.
Fig.~\ref{fig:framework} illustrates the complete pipeline.

\subsection{Relation State Construction and Flow Mapping}
\label{sec:relation}

\noindent\textbf{Feature projection and relation state.}
Raw intermediate features are ill-suited as a flow-map
state space: their absolute magnitudes are sensitive to
channel width and layer depth, making direct comparison
between teacher and student features of different
capacities unstable.
We instead define the distillation state on the
\emph{relation manifold}, where each state encodes the
pairwise structural geometry of a feature map rather
than its raw activations.
This representation is invariant to linear rescaling
of features and transfers more robustly across
architectures with mismatched
capacities~\cite{park2019rkd, tung2019similarity}.

Let $\mathcal{L} = \{2, 5, 8\}$ be the selected
distillation layers, covering approximately 20\%, 50\%,
and 80\% of network depth and capturing low-level
texture, mid-level structure, and high-level semantic
features respectively---matching the three-layer
configuration of FAKD~\cite{he2020fakd} and
DCKD~\cite{DCKD}.
For each layer $\ell \in \mathcal{L}$, teacher and
student features $f_t^\ell$, $f_s^\ell$ are projected
to a shared channel dimension $D$ via separate
$1{\times}1$ convolutions with batch normalisation
and ReLU, then spatially pooled to $p{\times}p$
($p{=}8$) via adaptive average pooling, yielding
$N{=}p^2{=}64$ tokens per layer.
This resolution balances representational expressiveness
against the $\mathcal{O}(N^2)$ state-space dimension of
the flow map operator; larger $p$ increases the relation
matrix dimension quadratically with diminishing returns,
as confirmed by the pooling-size ablation in
Table~\ref{tab:ablation_design}, where accuracy peaks at
$p{=}8$ while memory keeps growing as $\mathcal{O}(p^4)$.
After $\ell_2$ normalisation along the channel
dimension, the pairwise softmax affinity is computed as:
\begin{equation}
    R^\ell = \mathrm{softmax}\!\left(
        \frac{\tilde{X}^\ell (\tilde{X}^\ell)^\top}{\tau}
    \right) \in \mathbb{R}^{B \times N \times N},
    \quad \tau = 0.07,
\end{equation}
where the temperature $\tau{=}0.07$ follows the
contrastive learning convention~\cite{he2020momentum},
producing well-sharpened affinity distributions that
amplify structural differences between tokens.
The relation states are defined as:
\begin{equation}
    \begin{cases}
        z_0^\ell = \mathrm{vec}(R_s^\ell)
        & \text{(student, source)},\\[4pt]
        z_1^\ell = \mathrm{sg}\!\left(
            \mathrm{vec}(R_t^\ell)\right)
        & \text{(teacher, target)},
    \end{cases}
\end{equation}
where $\mathrm{sg}(\cdot)$ denotes stop-gradient.

\noindent\textbf{The relation manifold.}
Each row of $R^\ell$ is softmax-normalised and hence lies on the
probability simplex $\Delta^{N-1}$, so every relation state satisfies
$z^\ell \in (\Delta^{N-1})^N \subset \mathbb{R}^{N^2}$; this product of
simplices is the \emph{relation manifold} referred to throughout.
Because $\Delta^{N-1}$ is convex, linear interpolation between two
relation states stays within $(\Delta^{N-1})^N$: the bridge defined
below is intrinsic to the manifold rather than an artefact of its
Euclidean embedding, and every supervised intermediate state is itself
a valid relation configuration.

\noindent\textbf{Ground-truth bridge and time sampling.}
To avoid phantom state error accumulation, all intermediate
states used during training are drawn from the ground-truth
linear bridge between $z_0^\ell$ and $z_1^\ell$.
At each iteration we sample $t \sim \mathcal{U}[0, 0.8]$
and $\delta \sim \mathcal{U}[0, 1{-}t]$, setting
$s = t + \delta$ so that $t < s \leq 1$ is guaranteed,
ensuring a non-trivial interval for semigroup enforcement.
The two bridge states used throughout training are:
\begin{equation}
    \begin{aligned}
        z_t^\ell &= (1-t)\,z_0^\ell + t\,z_1^\ell, \\
        z_s^\ell &= (1-s)\,z_0^\ell + s\,z_1^\ell.
    \end{aligned}
\end{equation}

\noindent\textbf{Flow map operator.}
Unlike velocity-based methods that condition on a single
time index $t$, our operator takes \emph{both} source
and target times as independent inputs, which is essential
for modelling the asymmetric geometry of arbitrary
$(t \to s)$ transitions on the relation manifold.
Concretely, $t$ and $s$ are encoded via independent
sinusoidal positional
embeddings~\cite{vaswani2017attention} of dimension
$d{=}64$, and a learnable layer identity embedding
$e_\ell \in \mathbb{R}^d$ conditions the operator on
the distillation layer.
The full input to the 3-layer MLP with SiLU activations
and hidden dimension 256 is:
\begin{equation}
    \hat{z}_s = \mathcal{F}_\theta(h^\ell),
    \qquad
    h^\ell = \bigl[z_t^\ell \;\|\; e_t \;\|\; e_s
    \;\|\; e_\ell\bigr]
    \;\in\; \mathbb{R}^{N^2 + 3d}.
\end{equation}
The shallow MLP design is deliberate: a deeper operator
would risk overfitting the distillation trajectory and
increase training overhead, whereas a 3-layer network
with SiLU activations provides sufficient capacity to
model the nonlinear geometry of $(t \to s)$ transitions
on the relation manifold while adding fewer than 0.5M
parameters to the student pipeline.
A single $\mathcal{F}_\theta$ is shared across all
distillation layers, with $e_\ell$ providing
layer-specific conditioning, making FoRM a plug-in
module that requires no modification to the restoration
backbone.

\subsection{Structural Consistency and Training Objective}
\label{sec:objective}

\noindent\textbf{Safe semigroup consistency.}
\label{sec:sg}
A well-behaved flow map should satisfy the semigroup
property, ensuring compositional self-consistency
across arbitrary time decompositions:
\begin{equation}
    \mathcal{F}(z_t,\,t,\,1) \;=\;
    \mathcal{F}\!\bigl(\mathcal{F}(z_t,\,t,\,s),\;s,\;1\bigr),
    \quad \forall\; 0 \leq t < s \leq 1.
    \label{eq:semigroup}
\end{equation}
Enforcing Eq.~\eqref{eq:semigroup} by substituting the
predicted $\hat{z}_s$ as the right-branch input introduces
\emph{phantom state error
accumulation}~\cite{boffi2025consistency,
song2023consistency}: the predicted state drifts from
the true bridge, and the outer mapping compounds the
deviation.
We instead propose \emph{safe semigroup consistency}:
both branches are conditioned on ground-truth bridge
points, and stop-gradient is applied to the right branch
to prevent collapse:
\begin{equation}
    \mathcal{L}_{\mathrm{sg}}^\ell =
    \Bigl\|
        \mathcal{F}_\theta(z_t^\ell, t, 1, \ell)
        \;-\;
        \mathrm{sg}\!\left(
            \mathcal{F}_\theta(z_s^\ell, s, 1, \ell)
        \right)
    \Bigr\|_1.
\end{equation}

\noindent\textbf{Training objective.}
Two additional losses complete the framework:
\begin{equation}
    \begin{aligned}
        \mathcal{L}_{\mathrm{map}}^\ell
        &= \bigl\|\mathcal{F}_\theta(z_t^\ell, t, s, \ell)
           - z_s^\ell\bigr\|_2^2,
        \\[6pt]
        \mathcal{L}_{\mathrm{bd}}^\ell
        &= \bigl\|\mathcal{F}_\theta(z_t^\ell, t, 1, \ell)
           - z_1^\ell\bigr\|_2^2.
    \end{aligned}
\end{equation}
$\mathcal{L}_{\mathrm{map}}$ provides dense trajectory
supervision across arbitrary $(t{\to}s)$ pairs;
$\mathcal{L}_{\mathrm{bd}}$ anchors the operator to
the teacher target at $s{=}1$, without which
trajectories may be self-consistent but need not
converge to the teacher.
The total KD loss averages across layers under a
schedule weight $w(\rho)$:
\begin{equation}
    w(\rho) =
    \begin{cases}
        1 & \rho < \rho_0, \\[4pt]
        w_{\min} + (1 - w_{\min})
        \!\left(1 - \dfrac{\rho - \rho_0}{1 - \rho_0}\right)
        & \rho \geq \rho_0,
    \end{cases}
\end{equation}
where $\rho_0{=}0.6$ denotes the decay onset and
$w_{\min}{=}0.1$ the minimum weight.
$\rho_0$ is chosen so that the KD signal begins decaying
once the student has traversed the majority of the
relation gap; $w_{\min}$ retains a residual anchoring
signal to prevent manifold drift in the final training
phase.
The total loss is then:
\begin{equation}
    \mathcal{L}_{\mathrm{KD}} =
    \frac{1}{|\mathcal{L}|}\sum_\ell\!\left(
        w(\rho)\,\lambda_{\mathrm{map}}\,
        \mathcal{L}_{\mathrm{map}}^\ell
        + \lambda_{\mathrm{bd}}\,
        \mathcal{L}_{\mathrm{bd}}^\ell
        + \lambda_{\mathrm{sg}}\,
        \mathcal{L}_{\mathrm{sg}}^\ell
    \right),
\end{equation}
where $\lambda_{\mathrm{map}}{=}1.0$,
$\lambda_{\mathrm{bd}}{=}0.5$,
$\lambda_{\mathrm{sg}}{=}0.1$.
Note that $w(\rho)$ is applied exclusively to
$\mathcal{L}_{\mathrm{map}}$: the structural constraints
$\mathcal{L}_{\mathrm{bd}}$ and $\mathcal{L}_{\mathrm{sg}}$
remain at full weight throughout training, as decaying
them risks manifold drift and semigroup collapse in the
final training phase.
The final objective combines reconstruction and
distillation:
\begin{equation}
    \mathcal{L} = \mathcal{L}_{\mathrm{rec}}(\hat{y}, y)
    + \lambda_{\mathrm{KD}}\,\mathcal{L}_{\mathrm{KD}},
\end{equation}
where $\mathcal{L}_{\mathrm{rec}}$ is the $\ell_1$ pixel
loss.
Only the student, projectors, and map head are updated;
the teacher is frozen throughout.
The necessity of each component and their mutual
dependency are empirically confirmed in
Sec.~\ref{sec:ablation}.

\begin{figure*}[htp]
  \centering
  \includegraphics[width=\linewidth,height = 0.315\linewidth]{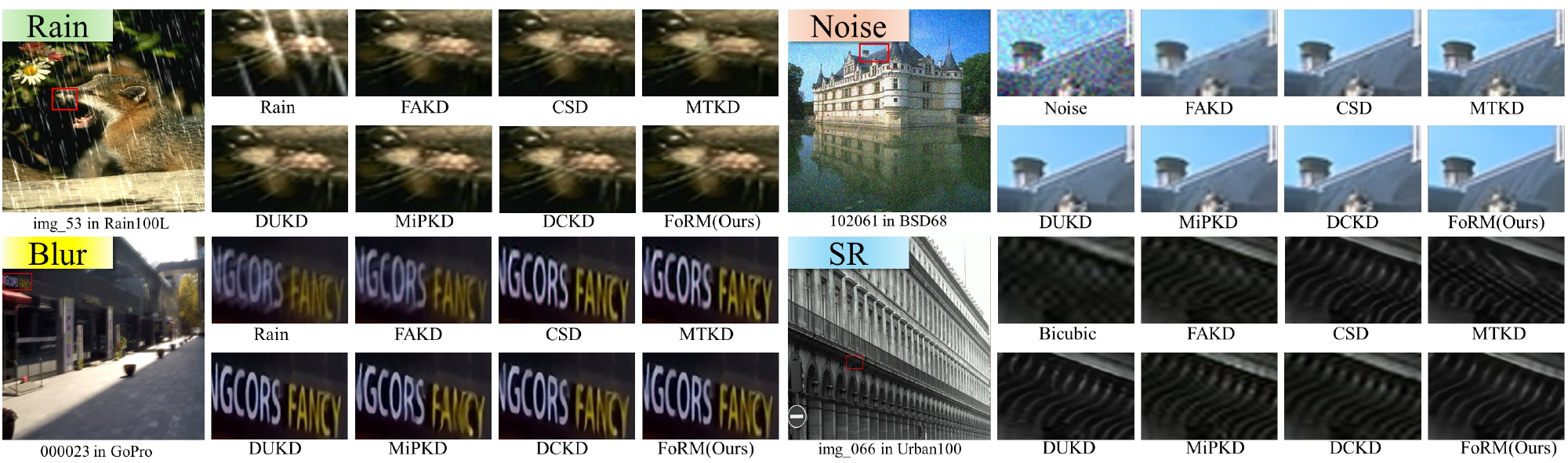}
  \Description{A grid of four quadrants, one per restoration task: deraining, denoising,
  deblurring and 4x super-resolution. Each quadrant shows the degraded input
  followed by cropped results from FAKD, CSD, MTKD, DUKD, MiPKD, DCKD and FoRM
  (Ours) on one scene, respectively a railing behind rain streaks, a building
  facade under noise, a shop sign bearing text under motion blur, and a brick
  building upsampled from a low-resolution crop. The FoRM crops show sharper
  edges and fewer artifacts than the competing methods.}
  \caption{
    \textbf{Qualitative comparison on four image restoration tasks.}
    For deraining (\textit{img\_53} in Rain100L), denoising (\textit{102061} in BSD68), 
    and deblurring (\textit{000023} in GoPro), we adopt Restormer~\cite{zamir2022restormer} 
    as the teacher--student backbone.
    For super-resolution $(\times4)$ (\textit{img\_066} in Urban100), 
    we adopt RCAN~\cite{zhang2018rcan}.
    Compared methods include FAKD~\cite{he2020fakd}, CSD~\cite{csd}, MTKD~\cite{mtkd}, 
    DUKD~\cite{DUKD}, MiPKD~\cite{MiPKD}, and DCKD~\cite{DCKD}.
    FoRM (Ours) consistently recovers finer structural details 
    with fewer artifacts across all four tasks.
  }
  \label{fig:qualitative_main}
\end{figure*}

\begin{figure*}[t]
  \centering
  \includegraphics[width=0.90\linewidth]{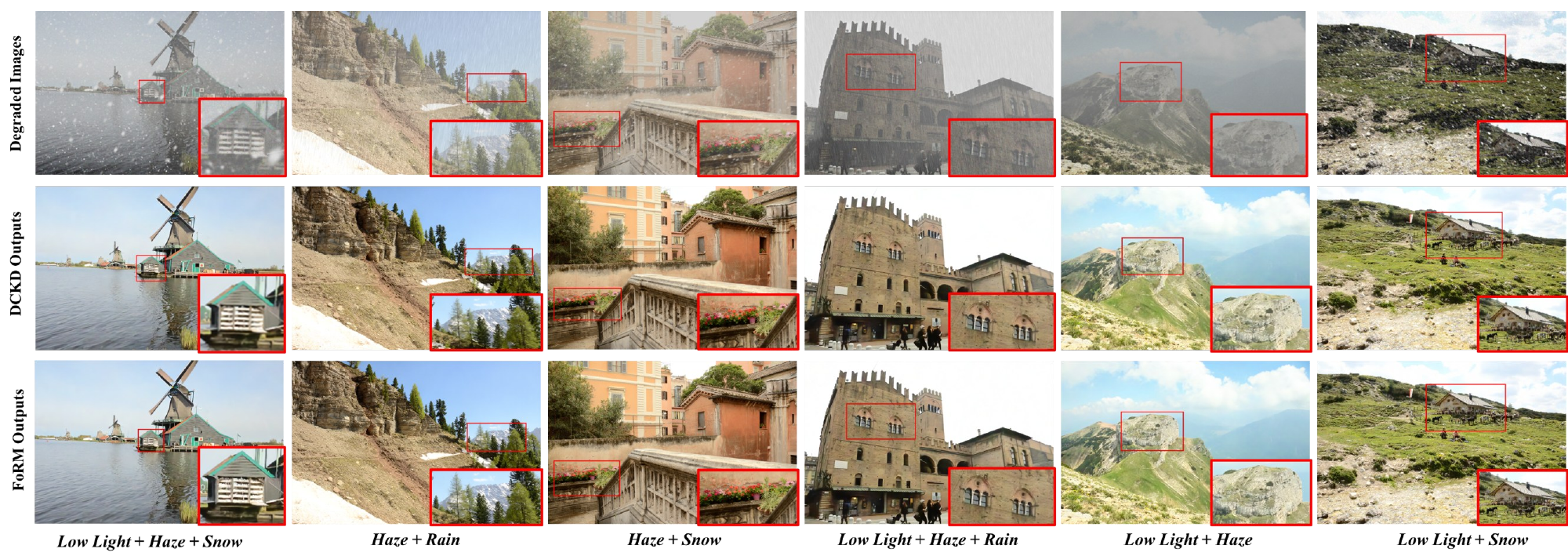}
  \Description{A grid of three rows and six columns. The rows are, from top to bottom, the
  degraded input, the DCKD output and the FoRM output. The columns correspond
  to six composite degradations: low light with haze and snow, haze with rain,
  haze with snow, low light with haze and rain, low light with haze, and low
  light with snow. The scenes are a windmill, a hillside, a city street, a
  castle, a hilltop fortress and a snowy mountain, and each panel carries a
  red inset box magnifying one detail. The FoRM row shows cleaner textures and
  more faithful colour than the DCKD row.}
  \caption{
    \textbf{Qualitative comparison on composite image degradation}
    (CDD-11~\cite{guo2024onerestore} dataset),
    using Restormer~\cite{zamir2022restormer} as the backbone.
    Each column corresponds to a different degradation combination:
    \textit{Low Light + Haze + Snow},
    \textit{Haze + Rain},
    \textit{Haze + Snow},
    \textit{Low Light + Haze + Rain},
    \textit{Low Light + Haze}, and
    \textit{Low Light + Snow}.
    From top to bottom: degraded input, DCKD~\cite{DCKD}, and FoRM (Ours).
    FoRM recovers cleaner textures and more faithful colors
    across all degradation combinations.
  }
  \label{fig:cdd11_compare}
  \vspace{-4mm}
\end{figure*}

\begin{figure}[t]
  \centering
  \includegraphics[width=\linewidth,height = 0.33\linewidth]{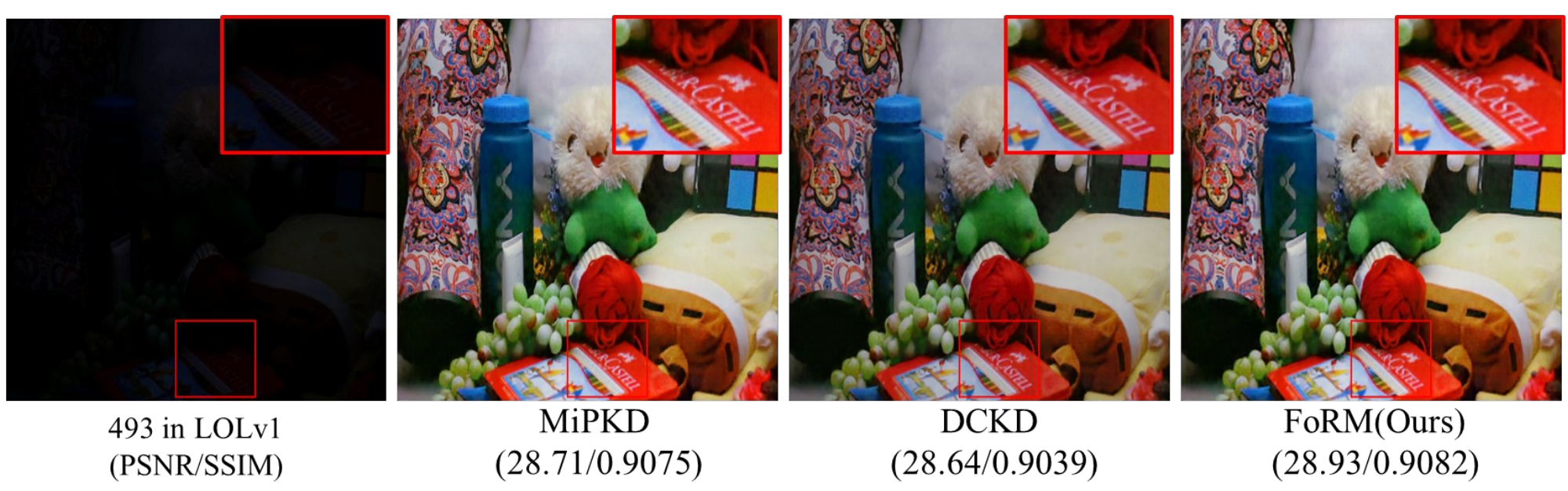}
  \Description{Four images side by side for a low-light indoor still life from LOLv1. From
  left to right: the almost entirely dark input, then the enhanced results of
  MiPKD at 28.71 dB PSNR and 0.9075 SSIM, of DCKD at 28.64 dB and 0.9039, and
  of FoRM (Ours) at 28.93 dB and 0.9082. Each result carries a red inset box
  magnifying the same region, in which the FoRM output shows the most legible
  package text and the least colour distortion.}
  \vspace{-5mm}
  \caption{
    \textbf{Qualitative comparison on low-light enhancement}
    (\textit{493} in LOLv1~\cite{wei2018deep}),
    using Restormer~\cite{zamir2022restormer} as the backbone.
    PSNR/SSIM scores are reported below each result.
  }
  \label{fig:lol_compare}
  \vspace{-6mm}
\end{figure}


\begin{table*}[t]
\centering
\caption{Quantitative comparison of knowledge distillation methods for
five-degradation all-in-one image restoration, using
Restormer~\cite{zamir2022restormer} as the teacher--student backbone.
Results are reported as PSNR\,(dB)\,$\uparrow$\,/\,SSIM\,$\uparrow$.
Denoising is evaluated at noise level $\sigma=25$.
The \textbf{Teacher} row denotes the full Restormer upper bound;
\textit{Scratch} denotes student training without distillation.}
\label{tab:main_restormer}
\renewcommand{\arraystretch}{1.12} 
\setlength{\tabcolsep}{8.5pt}    
\begin{tabular}{lcccccc}
\toprule
\multirow{2}{*}{\textbf{Method}} & \makecell{Dehazing \\ \small SOTS} & \makecell{Deraining \\ \small Rain100L} & \makecell{Denoising \\ \small BSD68} & \makecell{Deblurring \\ \small GoPro} & \makecell{Low-Light \\ \small LOLv1} & \makecell{\textbf{Average} \\ \textbf{Score}} \\
\midrule
\textbf{Teacher} & \textbf{24.09/0.927} & \textbf{34.81/0.962} & \textbf{31.49/0.884} & \textbf{27.22/0.829} & \textbf{20.41/0.806} & \textbf{27.60/0.881} \\
\addlinespace[0.5em] 
\midrule
Scratch & 21.83/0.891 & 32.14/0.941 & 30.67/0.864 & 25.31/0.798 & 18.76/0.778 & 25.74/0.854 \\
FAKD\cite{he2020fakd} & 23.66/0.901 & 34.11/0.943 & 31.01/0.864 & 26.61/0.793 & 19.54/0.782 & 26.99/0.857 \\
CSD\cite{csd} & 23.69/0.916 & 34.25/0.951 & 31.18/0.869 & 26.73/0.809 & 19.69/0.791 & 27.11/0.867 \\
MTKD\cite{mtkd} & 23.67/0.908 & 34.29/0.948 & 31.28/0.871 & 26.87/0.815 & 19.77/0.793 & 27.18/0.867 \\
MiPKD\cite{MiPKD} & 23.64/0.915 & 34.39/0.956 & 31.29/0.877 & 26.88/0.817 & 19.97/0.795 & 27.23/0.872 \\
DCKD\cite{DCKD} & 23.71/0.919 & 34.38/0.957 & 31.31/0.878 & 26.94/0.822 & 20.13/0.799 & 27.29/0.875 \\
\midrule
\rowcolor{gray!10} 
\textbf{FoRM (Ours)} & \textbf{23.78/0.921} & \textbf{34.42/0.959} & \textbf{31.39/0.881} & \textbf{26.95/0.823} & \textbf{20.18/0.801} & \textbf{27.34/0.877} \\
\bottomrule
\end{tabular}
\end{table*}

\begin{table*}[htbp]
    \centering
    \caption{\textbf{Quantitative comparison ($\times$4 SR)} of knowledge
    distillation methods across four backbone settings:
    SwinIR~\cite{liang2021swinir}, RCAN~\cite{zhang2018rcan},
    EDSR~\cite{lim2017edsr}, and a cross-architecture setting.
    All results are reported as PSNR\,(dB)\,/\,SSIM on Set5, Set14,
    BSD100, and Urban100.
    The \textbf{Teacher} row denotes the full-model upper bound;
    \textit{Scratch} denotes student training without distillation.
    The best and second-best results are \textbf{bolded} and
    \underline{underlined}, respectively.}
    \label{tab:beautified_results}
    \renewcommand{\arraystretch}{1.05}
    \resizebox{\textwidth}{!}{%
    \begin{tabular}{l cccc cccc}
        \toprule
        & \multicolumn{4}{c}{Method: \textbf{SwinIR}}
        & \multicolumn{4}{c}{Method: \textbf{RCAN}} \\
        \cmidrule(lr){2-5} \cmidrule(lr){6-9}
        Method & Set5 & Set14 & BSD100 & Urban100
               & Set5 & Set14 & BSD100 & Urban100 \\
        \midrule
        Teacher
        & 32.72/0.9021 & 28.94/0.7914 & 27.83/0.7459 & 27.07/0.8164
        & 32.63/0.9002 & 28.87/0.7889 & 27.77/0.7436 & 26.82/0.8087 \\
        Scratch
        & 32.31/0.8955 & 28.67/0.7833 & 27.61/0.7379 & 26.15/0.7884
        & 32.38/0.8971 & 28.69/0.7842 & 27.63/0.7379 & 26.36/0.7947 \\
        Logits~\cite{logits}
        & 32.27/0.8954 & 28.67/0.7833 & 27.62/0.7380 & 26.15/0.7887
        & 32.45/0.8980 & 28.76/0.7860 & 27.67/0.7400 & 26.49/0.7982 \\
        FAKD~\cite{he2020fakd}
        & 32.22/0.8950 & 28.65/0.7831 & 27.61/0.7379 & 26.09/0.7870
        & 32.46/0.8980 & 28.77/0.7860 & 27.68/0.7409 & 26.50/0.7980 \\
        MiPKD~\cite{MiPKD}
        & 32.39/0.8971 & 28.76/0.7854 & 27.68/0.7403 & 26.37/0.7956
        & 32.46/0.8982 & 28.77/0.7860 & 27.69/0.7412 & 26.55/0.7998 \\
        \rowcolor{grayhighlight}
        DCKD~\cite{DCKD}
        & \underline{32.49/0.8991} & \underline{28.82/0.7877}
        & \underline{27.72/0.7422} & \underline{26.53/0.8007}
        & \underline{32.56/0.8995} & \underline{28.82/0.7877}
        & \underline{27.73/0.7423} & \underline{26.69/0.8041} \\
        \rowcolor{grayhighlight}
        \textbf{FoRM}
        & \textbf{32.51/0.8992} & \textbf{28.88/0.7890}
        & \textbf{27.74/0.7430} & \textbf{26.62/0.8032}
        & \textbf{32.58/0.8996} & \textbf{28.86/0.7885}
        & \textbf{27.74/0.7425} & \textbf{26.74/0.8054} \\
        \midrule
        & \multicolumn{4}{c}{Method: \textbf{EDSR}}
        & \multicolumn{4}{c}{Method: \textbf{Cross-Architecture}} \\
        \cmidrule(lr){2-5} \cmidrule(lr){6-9}
        Teacher
        & 32.60/0.8998 & 28.80/0.7876 & 27.71/0.7420 & 26.66/0.8010
        & 32.63/0.9002 & 28.87/0.7889 & 27.77/0.7436 & 26.82/0.8087 \\
        Scratch
        & 32.25/0.8940 & 28.60/0.7810 & 27.55/0.7350 & 26.05/0.7850
        & 32.35/0.8965 & 28.70/0.7845 & 27.65/0.7380 & 26.25/0.7920 \\
        Logits\cite{logits}
        & 32.30/0.8950 & 28.65/0.7825 & 27.58/0.7365 & 26.10/0.7865
        & 32.42/0.8975 & 28.71/0.7858 & 27.68/0.7405 & 26.40/0.7960 \\
        FAKD\cite{he2020fakd}
        & 32.32/0.8955 & 28.68/0.7830 & 27.60/0.7370 & 26.12/0.7870
        & 32.44/0.8978 & 28.74/0.7863 & 27.65/0.7404 & 26.45/0.7975 \\
        MiPKD\cite{MiPKD}
        & 32.38/0.8968 & 28.74/0.7850 & 27.65/0.7395 & 26.30/0.7935
        & 32.48/0.8985 & 28.76/0.7867 & 27.67/0.7411 & 26.58/0.8005 \\
        \rowcolor{grayhighlight}
        DCKD\cite{DCKD}
        & \underline{32.44/0.8985} & \underline{28.79/0.7868}
        & \underline{27.69/0.7412} & \underline{26.48/0.7985}
        & \underline{32.54/0.8992} & \underline{28.81/0.7872}
        & \underline{27.71/0.7419} & \underline{26.68/0.8035} \\
        \rowcolor{grayhighlight}
        \textbf{FoRM}
        & \textbf{32.47/0.8990} & \textbf{28.82/0.7872}
        & \textbf{27.70/0.7418} & \textbf{26.55/0.8002}
        & \textbf{32.57/0.8995} & \textbf{28.84/0.7880}
        & \textbf{27.72/0.7424} & \textbf{26.72/0.8050} \\
        \bottomrule
    \end{tabular}}
\end{table*}

\section{Experiments}
\label{sec:experiments}
We evaluate FoRM across three dimensions:
\emph{reconstruction quality} (PSNR/SSIM on standard
benchmarks across five restoration tasks),
\emph{training stability} (late-stage variance and
convergence behaviour), and \emph{component necessity}
(ablation of the three structural losses).
We additionally report a diagnostic analysis of the
optimal transport baseline in the Appendix.

\subsection{Experimental Setup}
\label{sec:setup}
{\sloppy
\noindent\textbf{Datasets.}
For \emph{image restoration} we follow the all-in-one protocol
of~\cite{zamir2022restormer, bendbasics} on five tasks---dehazing
(SOTS~\cite{li2019benchmarking}), deraining
(Rain100L~\cite{yang2017deep}), denoising
(BSD68~\cite{martin2001database}, $\sigma{=}25$), deblurring
(GoPro~\cite{nah2017deep}) and low-light enhancement
(LOLv1~\cite{wei2018deep})---plus composite degradation on
CDD-11~\cite{guo2024onerestore}.
For \emph{super-resolution} ($\times$4), models are
trained on DIV2K~\cite{agustsson2017ntire} (800 images)
and evaluated on Set5~\cite{bevilacqua2012low},
Set14~\cite{zeyde2010single},
B100~\cite{martin2001database},
and Urban100~\cite{huang2015single}.

\noindent\textbf{Teacher and student architectures.}
For restoration the teacher is Restormer~\cite{zamir2022restormer}
in its default full configuration and the student a reduced-depth
variant of the same architecture. For super-resolution we evaluate
four backbone settings: RCAN~\cite{zhang2018rcan},
SwinIR~\cite{liang2021swinir}, EDSR~\cite{lim2017edsr}, and a
cross-architecture setting pairing an RCAN teacher with a plain
residual student without channel attention. 
For RCAN, the teacher uses 10 residual groups,
20 residual blocks, and 64 channels (15.59M parameters,
1044.03 GFLOPs on a $256{\times}256$ input), and the
student uses the same architecture with 6 residual
blocks (5.17M parameters, 366.98 GFLOPs).
Distillation is applied at layer indices
$\mathcal{L} = \{2, 5, 8\}$, covering approximately
20\%, 50\%, and 80\% of network depth.

\noindent\textbf{Training details.}
All models are trained with the Adam
optimiser~\cite{kingma2014adam}
($\beta_1{=}0.9$, $\beta_2{=}0.999$),
an initial learning rate of $1{\times}10^{-4}$ halved
at epoch 250, batch size 16, and $96{\times}96$
input patches with random horizontal/vertical flipping
and rotation.
The total training budget is 500 epochs;
$\lambda_{\mathrm{KD}}{=}100$.
For stability experiments, each configuration is
run with 3 independent random seeds.

\noindent\textbf{Evaluation protocol.}
For super-resolution, PSNR and SSIM are computed on
the Y channel of the YCbCr colour space (BT.601
coefficients) with a border shave of 4 pixels,
matching the evaluation convention of
RCAN~\cite{zhang2018rcan}.
For restoration tasks, PSNR and SSIM are computed
on the full RGB image following the protocol
of~\cite{zamir2022restormer}.
We report \emph{final-epoch} PSNR as the primary
metric (epoch 500) to avoid cherry-picking across
runs; best-epoch PSNR is reported in parentheses
where noted.
Training stability is quantified by the standard
deviation of PSNR over epochs 451--500
($\sigma_{\mathrm{late}}$) and, where seeds are
available, the cross-seed standard deviation of
final PSNR ($\sigma_{\mathrm{seed}}$).
}

\subsection{Comparison with State-of-the-Art Methods}

\noindent\textbf{Qualitative Evaluation.}
Figures~\ref{fig:qualitative_main}--\ref{fig:lol_compare} compare all five
restoration tasks. On single-degradation tasks
(Figure~\ref{fig:qualitative_main}) FoRM outperforms competing KD methods
across deraining, denoising, deblurring and super-resolution ($\times4$):
FAKD and CSD leave residual artifacts and blurry textures, while DCKD, the
strongest baseline, still over-smooths fine-grained regions such as roof
edges and text boundaries. FoRM recovers sharper structural detail with
fewer artifacts throughout, reflecting superior relational alignment with
the teacher.

\begin{figure*}[htp]
  \centering
  \includegraphics[width=0.92\linewidth]{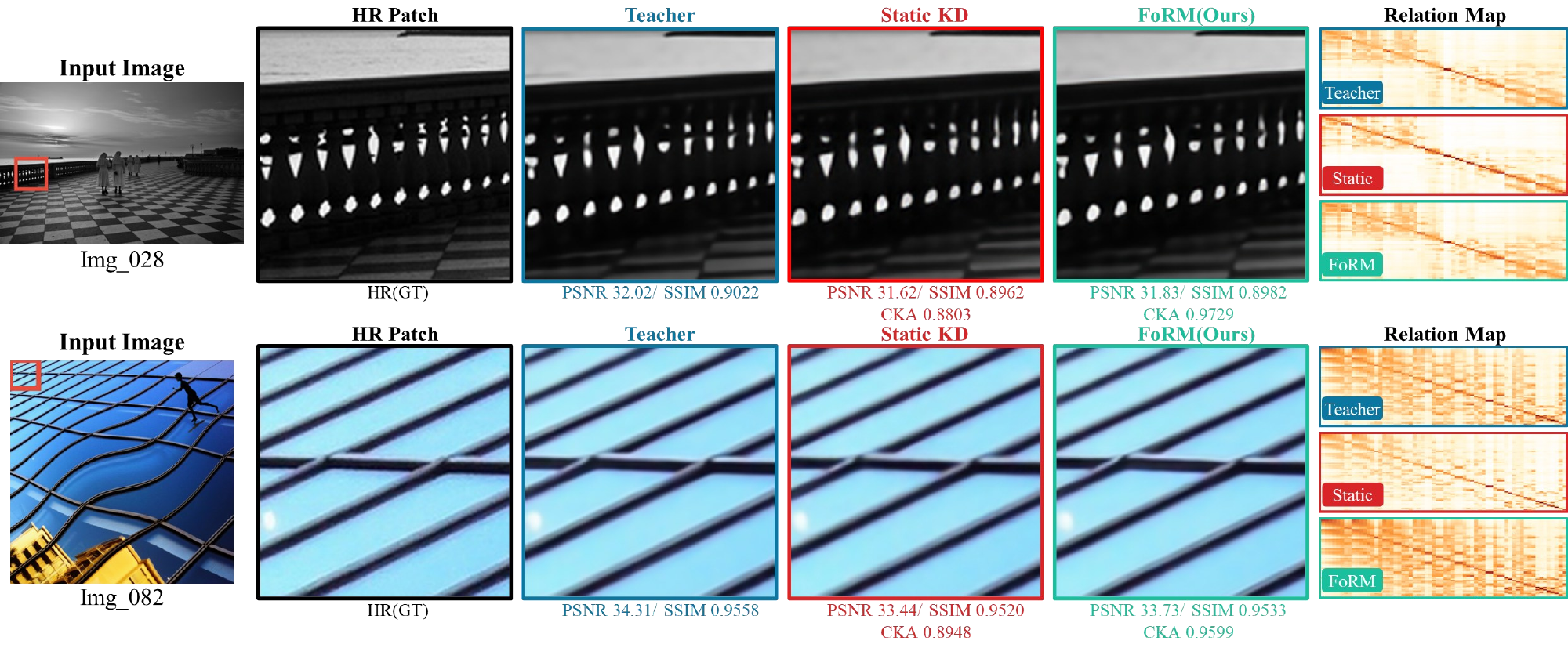}
  \Description{Two rows, one per Urban100 example. Each row shows, from left to right, the
  input image, the high-resolution ground-truth patch, and the super-resolved
  patches produced by the Teacher, by Static KD and by FoRM (Ours), followed
  by a column of three relation maps rendered as heat maps. The first row is a
  scene of repeated bright lights, the second a facade of diagonal stripes.
  PSNR, SSIM and CKA values are printed beneath each patch: FoRM attains CKA
  0.9729 against 0.8803 for Static KD on the first example, and 0.9599 against
  0.8948 on the second.}
  \vspace{-6mm}
  \caption{
    \textbf{Qualitative comparison on Urban100 ($\times$4).}
    For each example, we show the HR patch (ground truth), 
    super-resolved patches from the Teacher, Static KD, and FoRM (Ours),
    along with their corresponding relation maps.
    FoRM achieves higher CKA with the Teacher's relation map
    (0.9729 vs.\ 0.8803 on \textit{img\_028};
     0.9599 vs.\ 0.8948 on \textit{img\_082}),
    indicating better relational structure alignment,
    which correlates with improved PSNR and SSIM.
  }
  \label{fig:qualitative}
  \vspace{-3mm}
\end{figure*}

For composite and low-light degradation
(Figures~\ref{fig:cdd11_compare} and~\ref{fig:lol_compare}),
FoRM generalizes effectively to more challenging conditions.
On CDD-11~\cite{guo2024onerestore} under six multi-degradation combinations,
DCKD exhibits residual haze and color shifts, whereas FoRM produces
consistently cleaner restorations.
On LOLv1 low-light enhancement, FoRM achieves the highest
PSNR of 28.93\,dB and SSIM of 0.9082, surpassing MiPKD (28.71/0.9075)
and DCKD (28.64/0.9039), with finer texture recovery in both
shadow and highlight regions.

\noindent\textbf{Quantitative Evaluation.}
Table~\ref{tab:main_restormer} reports results on five-degradation all-in-one
restoration with Restormer as the backbone.
FoRM achieves the best average score of 27.34\,dB, outperforming
DCKD (27.29\,dB\,/\allowbreak\,0.875) on every individual task, with the largest gains
on low-light enhancement and dehazing.
Table~\ref{tab:beautified_results} further evaluates $\times$4 SR across
SwinIR, RCAN, EDSR and a cross-architecture setting; FoRM ranks first on
all four benchmarks under every configuration, most notably on Urban100
whose structured repetitive textures benefit most from relation map
alignment. Improving over DCKD in both homogeneous and cross-architecture
settings shows that flow-map distillation supplies stronger and more
generalizable intermediate supervision than static feature alignment.

\begin{figure}[htp]
  \centering
  \includegraphics[width=0.95\linewidth]{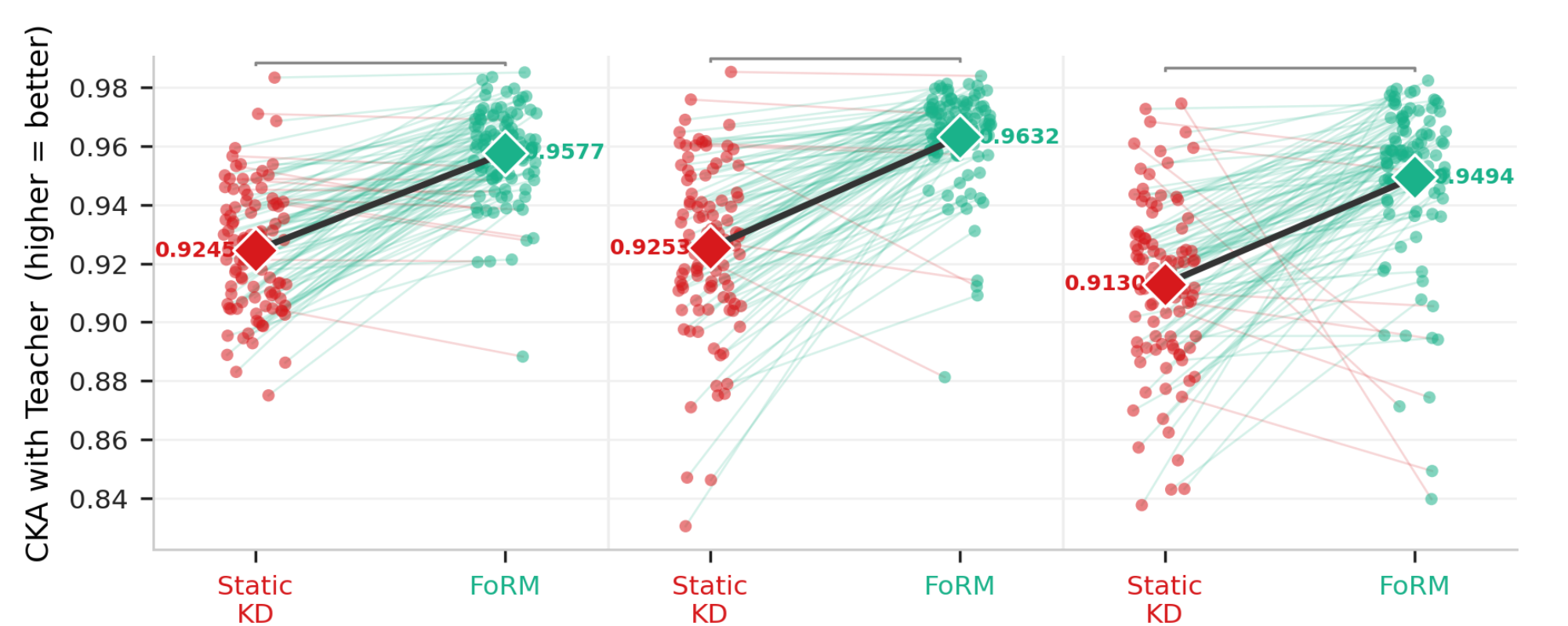}
  \Description{Three paired scatter plots side by side, one per distillation layer. The
  vertical axis is CKA with the teacher, where higher is better, spanning
  roughly 0.84 to 0.98. Within each panel the left cluster of red points is
  Static KD and the right cluster of green points is FoRM, thin lines connect
  the two scores belonging to the same image, and a large diamond marks the
  mean. The means rise from 0.9243 to 0.9577, from 0.9252 to 0.9632, and from
  0.9130 to 0.9494, with a significance bar spanning each pair.}
  \caption{
    \textbf{Per-image CKA with Teacher across three layers (Urban100, $\times$4).}
    Each point represents one image; connected lines link the same image's
    Static KD and FoRM scores.
    Large diamonds denote the mean CKA.
    FoRM consistently achieves higher CKA with the Teacher across all layers
    (Wilcoxon signed-rank test, $p < 0.001$).
  }
  \label{fig:cka_paired}
  \vspace{-5mm}
\end{figure}

\noindent\textbf{Relational Alignment Analysis}
To examine whether the qualitative improvements in 
Fig.~\ref{fig:qualitative} reflect a consistent trend 
across the full Urban100 benchmark, Fig.~\ref{fig:cka_paired} 
reports per-image CKA with the Teacher for all 100 images 
across three representative residual groups. FoRM consistently 
achieves higher CKA at every layer---improving mean CKA from 
0.9243 to 0.9577 (layer~1), 0.9253 to 0.9632 (layer~2), 
and 0.9130 to 0.9494 (layer~3)---with all differences 
confirmed by the Wilcoxon signed-rank test ($p < 0.001$). 
This result demonstrates that FoRM's PSNR gains are 
underpinned by systematically superior relational structure 
alignment with the Teacher, rather than isolated per-image 
improvements.

\subsection{Training Stability Analysis}
\label{sec:stability}

A key advantage of FoRM is that the semigroup constraint
stabilizes training relative to static relation distillation and
unconstrained flow matching. Three configurations share all
hyperparameters ($\lambda_{\mathrm{KD}}$=100, learning rate schedule,
500-epoch budget) and differ only in the distillation objective:
(i)~\textit{Static Rel.\ KD}, regressing the student relation map
onto the teacher target via an $\ell_1$ loss with no flow-map
operator;
(ii)~\textit{Naive FM}, replacing that static target with a
constant-velocity field $\mathbf{u}^* = z_1 - z_0$---equivalent to
dropping all three structured losses ($\mathcal{L}_{\mathrm{map}}$,
$\mathcal{L}_{\mathrm{bd}}$, $\mathcal{L}_{\mathrm{sg}}$) and
regressing only the endpoint displacement, which isolates
continuous-time supervision without compositional consistency; and
(iii)~\textit{FoRM}, the full system.
Table~\ref{tab:stability} reports two stability metrics on
Set5 $\times$4: $\sigma_{\mathrm{late}}$, the PSNR standard deviation
over epochs 451--500 (the post-decay window where the KD signal is
weakest and residual oscillation most visible), and
$\sigma_{\mathrm{seed}}$, the standard deviation of final PSNR over
3 independent random seeds.

\begin{table}[htp]
\centering
\small
\caption{\textbf{Left:} training stability on Set5 $\times$4.
$\sigma_{\mathrm{late}}$: std of validation PSNR over epochs 451--500
(single seed); $\sigma_{\mathrm{seed}}$: std of final PSNR across 3
seeds. The three recent flow-distillation objectives are adapted to
KD by replacing $(\mathcal{L}_{\mathrm{map}},\mathcal{L}_{\mathrm{bd}},
\mathcal{L}_{\mathrm{sg}})$ on the same state $z$, leaving the
reconstruction loss and epoch budget unchanged.
\textbf{Right:} training cost per iteration (A100, identical
protocol); $\mathcal{F}_\theta$ is discarded after training, so
inference cost is unchanged.}
\label{tab:stability}
\renewcommand{\arraystretch}{1.12}
\begin{minipage}[t]{0.585\linewidth}
\centering\footnotesize
\setlength{\tabcolsep}{2.5pt}
\begin{tabular}{lccc}
\toprule
Method & PSNR
  & $\sigma_{\mathrm{late}}$ $\downarrow$
  & $\sigma_{\mathrm{seed}}$ $\downarrow$ \\
\midrule
Static Rel.\ KD  & 31.83 & 0.0312 & 0.0241 \\
Naive FM         & 31.73 & 0.0680 & 0.0318 \\
\midrule
CTM~\cite{kim2024ctm}          & 31.90 & 0.0264 & -- \\
AYF~\cite{sabour2025alignflow} & 31.98 & 0.0221 & -- \\
Stable Vel.~\cite{lee2026stable} & 32.04 & 0.0193 & -- \\
\midrule
\best{FoRM (Ours)} & \best{32.13} & \best{0.0150} & \best{0.0112} \\
\bottomrule
\end{tabular}
\end{minipage}\hfill
\begin{minipage}[t]{0.395\linewidth}
\centering\footnotesize
\setlength{\tabcolsep}{2pt}
\begin{tabular}{lccc}
\toprule
Backbone & DCKD & FoRM & $\Delta$ \\
\midrule
RCAN      & 0.40 & 0.42 & $+$5\% \\
SwinIR    & 0.68 & 0.71 & $+$4\% \\
Restormer & 1.00 & 1.05 & $+$5\% \\
\midrule
\multicolumn{4}{@{}l@{}}{\scriptsize s/iter; scratch (RCAN) 0.38,} \\
\multicolumn{4}{@{}l@{}}{\scriptsize MiPKD 0.41; mem $+$0.4\,GB} \\
\bottomrule
\end{tabular}
\end{minipage}
\end{table}

FoRM reduces $\sigma_{\mathrm{late}}$ by \textbf{52\%} relative to
Static Rel.\ KD and by \textbf{78\%} relative to Naive FM, and also
attains the highest PSNR (32.13\,dB): stability comes without a
quality trade-off.
The comparison extends to three recent flow-distillation objectives
adapted to KD, over which FoRM stays ahead by 0.09--0.23\,dB at
1.3--1.8$\times$ lower $\sigma_{\mathrm{late}}$; notably even Stable
Vel.~\cite{lee2026stable}, which explicitly targets variance
reduction, does not match the safe semigroup design.
This margin costs $+$4--5\% training time and $+$0.4\,GB memory, and
nothing at inference: $\mathcal{F}_\theta$ is discarded once training
ends, so the deployed student is identical to the baseline.

\subsection{Ablation Study}
\label{sec:ablation}

\begin{table}[htp]
    \centering
    \small
    \renewcommand{\arraystretch}{1.15}
    \caption{Component ablation on Set5 $\times$4. B0: raw-feat.\
    $\ell_2$ KD; B1: static relation KD; A0: naive velocity FM;
    A3$^*$: $\mathcal{L}_{\mathrm{sg}}$ without $\mathcal{L}_{\mathrm{bd}}$;
    $\sigma$: PSNR std over ep\,451--500.}
    \label{tab:ablation_component}
    \begin{tabularx}{\columnwidth}{@{} l *{3}{>{\centering\arraybackslash}X}
                                       *{2}{>{\centering\arraybackslash}X}
                                       >{\centering\arraybackslash}X @{}}
        \toprule
        & \multicolumn{3}{c}{Components}
          & \multicolumn{3}{c}{Metrics} \\
        \cmidrule(lr){2-4} \cmidrule(lr){5-7}
        Config
        & $\mathcal{L}_{\mathrm{map}}$
        & $\mathcal{L}_{\mathrm{bd}}$
        & $\mathcal{L}_{\mathrm{sg}}$
        & PSNR $\uparrow$ & SSIM $\uparrow$ & $\sigma$ $\downarrow$ \\
        \midrule
        B0 & \xmark & \xmark & \xmark & 31.61 & 0.8882 & 0.0389 \\
        B1 & \xmark & \xmark & \xmark & 31.83 & 0.8915 & 0.0312 \\
        \midrule
        A0 & \xmark & \xmark & \xmark & 31.73 & 0.8906 & 0.0680 \\
        A1 & \cmark & \xmark & \xmark & 31.89 & 0.8918 & 0.0141 \\
        A2 & \cmark & \cmark & \xmark & 32.01 & 0.8927 & 0.0134 \\
        A3$^*$ & \cmark & \xmark & \cmark & 31.94 & 0.8921 & 0.0391 \\
        \rowcolor{bestgray}
        A3 & \cmark & \cmark & \cmark
           & \textbf{32.13} & \textbf{0.8938} & \textbf{0.0098} \\
        \bottomrule
    \end{tabularx}
\end{table}

\noindent\textbf{Component necessity.}
Table~\ref{tab:ablation_component} ablates two questions
in sequence: whether the relation-manifold representation
is necessary over raw-feature Euclidean alignment, and
how each structural loss contributes within the flow-map
framework.
All configurations share identical hyperparameters and
differ only in representation space and distillation objective.

{\sloppy
B0 applies $\ell_2$ matching on raw activations
(FitNets-style~\cite{romero2014fitnets}); B1 (Static Rel.\ KD)
operates on the softmax-normalised relation matrix.
The +0.22\,dB gain and $\sigma_{\mathrm{late}}$ reduction
(0.0389\allowbreak$\,{\to}\,$0.0312) confirm that the relation manifold
provides a more stable distillation target by encoding
structural geometry invariant to linear feature
rescaling~\cite{park2019rkd}.
Notably, naively replacing B1 with a constant-velocity
field (A0) \emph{decreases} PSNR by 0.10\,dB and more than
doubles $\sigma_{\mathrm{late}}$, confirming that
continuous-time supervision alone is insufficient without
compositional structure~\cite{lee2026stable}.

Within the flow-map framework, each component contributes
monotonically. The asymmetric $(t,s)$-conditioned operator
(A0$\to$A1, +0.16\,dB) provides richer trajectory supervision than
constant velocity; endpoint anchoring (A1\allowbreak$\,{\to}\,$A2,
+0.12\,dB) supplies the absolute teacher reference, without which
trajectories are self-consistent yet need not converge to the
teacher.
A3$^*$ verifies that $\mathcal{L}_{\mathrm{bd}}$ is a necessary
precondition for $\mathcal{L}_{\mathrm{sg}}$~\cite{boffi2025consistency}:
applying $\mathcal{L}_{\mathrm{sg}}$ alone gives
$\sigma_{\mathrm{late}}{=}0.0391$, worse than even B1---without an
anchor at $s{=}1$ the semigroup loss enforces consistency around a
drifting target, amplifying rather than dampening oscillation.
Pairing both losses (A3) yields the largest $\sigma_{\mathrm{late}}$
reduction ($0.0134{\to}0.0098$, $-$27\%) plus a further +0.12\,dB.
}

\noindent\looseness=-1\textbf{Design choices.}
Table~\ref{tab:ablation_design} examines four secondary design
decisions shaping the efficiency and stability of FoRM.
Accuracy peaks at pooling size $p{=}8$: smaller grids under-resolve
the relation structure, while $p{=}12$ regresses slightly despite
$5\times$ the memory---the diminishing return noted in
Sec.~\ref{sec:method}.
Restricting $t \in [0, 0.8]$ outperforms the full range
($+$0.09\,dB): sampled near 1, the interval $1{-}t$ becomes
negligible and the semigroup constraint degenerates to a trivial
identity, contributing noise rather than structure to the
consistency loss.
The schedule weight $w(\rho)$ is critical: without late-stage
decay, the KD signal over-constrains the student after the relation
gap has substantially closed, inflating $\sigma_{\mathrm{late}}$ by
34\% and suppressing fine-detail recovery late in training.
Finally, sharing $\mathcal{F}_\theta$ across layers with a learnable
layer-identity embedding $e^\ell$ matches or slightly exceeds
per-layer independent heads ($+$0.04\,dB) at half the distillation
parameter overhead, suggesting cross-layer weight sharing acts as a
mild regulariser on the flow-map operator.

\begin{table}[htp]
    \centering
    \small
    \setlength{\tabcolsep}{0pt}
    \renewcommand{\arraystretch}{1.18}
    \caption{Design choice ablation (500 epochs). Pooling size
    $p$ additionally reports the extra memory $\Delta$M over the
    $p{=}4$ setting, which grows as $\mathcal{O}(p^4)$.}
    \label{tab:ablation_design}
    \begin{tabular*}{\linewidth}{@{\extracolsep{\fill}} ll ccc @{}}
        \toprule
        Factor & Variant & PSNR $\uparrow$ & $\sigma_{\mathrm{late}}$ $\downarrow$
        & $\Delta$M \\
        \midrule
        \multirow{4}{*}{Pooling size $p$}
        & $p{=}4$                        & 31.97 & 0.0183 & 0.03\,GB \\
        & $p{=}6$                        & 32.08 & 0.0162 & 0.13\,GB \\
        & \cellcolor{bestgray}$p{=}8$
        & \best{32.13} & \best{0.0150} & \cellcolor{bestgray}0.40\,GB \\
        & $p{=}12$                       & 32.11 & 0.0168 & 2.00\,GB \\
        \midrule
        \multirow{2}{*}{Time range}
        & $t \sim \mathcal{U}[0,1]$      & 32.04 & 0.0143 & -- \\
        & \cellcolor{bestgray}$t \sim \mathcal{U}[0,0.8]$
        & \best{32.13} & \best{0.0150} & \cellcolor{bestgray}-- \\
        \midrule
        \multirow{2}{*}{Schedule $w(\rho)$}
        & Off (constant $w{=}1$)         & 31.98 & 0.0201 & -- \\
        & \cellcolor{bestgray}On (decay at $\rho{=}0.6$)
        & \best{32.13} & \best{0.0150} & \cellcolor{bestgray}-- \\
        \midrule
        \multirow{2}{*}{Map head}
        & Per-layer (independent)        & 32.09 & 0.0158 & -- \\
        & \cellcolor{bestgray}Shared $+$ layer embedding
        & \best{32.13} & \best{0.0150} & \cellcolor{bestgray}-- \\
        \bottomrule
    \end{tabular*}
    \vspace{-4mm}
\end{table}

\begin{table}[htbp]
    \centering
    \small
    \renewcommand{\arraystretch}{1.12}
    \setlength{\tabcolsep}{8pt}
    \caption{Safe vs.\ naive recursive semigroup on
    Set5 $\times$4. ``sg'': stop-gradient on right branch.}
    \label{tab:ablation_sg}
    \begin{tabularx}{\columnwidth}{@{} X cc @{}}
        \toprule
        Semigroup variant & PSNR & $\sigma_{\mathrm{late}}$ \\
        \midrule
        Naive (recursive $\hat{z}_s$, no sg)  & 31.87 & 0.0312 \\
        Naive (recursive $\hat{z}_s$, $+$sg)  & 31.94 & 0.0228 \\
        \rowcolor{bestgray}
        \textbf{Safe (GT $z_s$, $+$sg)}
        & \textbf{32.13} & \textbf{0.0098} \\
        \bottomrule
    \end{tabularx}
    \vspace{-4mm}
\end{table}

\noindent\looseness=-1\textbf{Safe vs.\ naive semigroup.}
Table~\ref{tab:ablation_sg} isolates the ``safe'' design by comparing
three variants differing in how the right branch of
$\mathcal{L}_{\mathrm{sg}}$ is conditioned. The first naive variant
feeds the model-predicted $\hat{z}_s$ back into that branch without
stop-gradient, letting gradient flow through both paths; the second
adds stop-gradient but still uses the predicted state as input.
Both underperform substantially: the better one trails Full FoRM by
$-$0.19\,dB with $2.3\times$ higher $\sigma_{\mathrm{late}}$, and the
no-stop-gradient variant is worst on both ($-$0.26\,dB, $3.2\times$).
This directly quantifies the phantom state error
accumulation described in Sec.~\ref{sec:sg}: predicted
states drift from the true bridge, and the outer mapping
compounds the deviation regardless of whether gradients
are blocked.
The safe variant eliminates this by conditioning both
branches on ground-truth bridge points, confirming that
the design choice is essential rather than incidental.

\section{Conclusion}
\label{sec:conclusion}
\looseness=-2
\looseness=-2
We presented FoRM, which recasts relation-based knowledge transfer as
continuous flow mapping on the relation manifold. A flow map operator
supervised by map distillation, endpoint anchoring and safe semigroup
consistency replaces static endpoint alignment with trajectory-level
supervision. Across five restoration tasks FoRM improves consistently
over static relation distillation, and ablations validate every
component as well as the phantom-state error mechanism behind the safe
semigroup design.

\bibliographystyle{ACM-Reference-Format}
\balance
\bibliography{sample-base}










\end{document}